\documentclass[final]{main}

\usepackage{times}
\usepackage{epsfig}
\usepackage{graphicx}
\usepackage{amsmath}
\usepackage{amssymb}
\usepackage{times}
\usepackage{epsfig}
\usepackage{makecell}
\usepackage{subcaption}
\usepackage{caption}
\usepackage{float}

\usepackage[pagebackref=true,breaklinks=true,colorlinks,bookmarks=false]{hyperref}
\usepackage[dvipsnames]{xcolor}
\definecolor{aqua}{RGB}{102, 178, 255}
\definecolor{blue}{RGB}{0, 0, 255}

\def\cvprPaperID{10276} 
\def\confYear{CVPR 2021}

\begin{document}

\title{GCNNMatch: Graph Convolutional Neural Networks for Multi-Object Tracking via Sinkhorn Normalization}

\author{Ioannis Papakis\\
Virginia Tech\\
Blacksburg, VA, USA\\
{\tt\small iopapakis@vt.edu}
\and
Abhijit Sarkar\\
Virginia Tech Transportation Insitute\\
Blacksburg, VA, USA\\
{\tt\small asarkar1@vt.edu}
\and
Anuj Karpatne\\
Virginia Tech\\
Blacksburg, VA, USA\\
{\tt\small karpatne@vt.edu}
}

\maketitle

\begin{abstract}
This paper proposes a novel method for online Multi-Object Tracking (MOT) using Graph Convolutional Neural Network (GCNN) based feature extraction and end-to-end feature matching for object association. The Graph based approach incorporates both appearance and geometry of objects at past frames as well as the current frame into the task of feature learning. This new paradigm enables the network to leverage the ``context'' information of the geometry of objects and allows us to model the interactions among the features of multiple objects. Another central innovation of our proposed framework is the use of the Sinkhorn algorithm for end-to-end learning of the associations among objects during model training. The network is trained to predict object associations by taking into account constraints specific to the MOT task. Experimental results demonstrate the efficacy of the proposed approach in achieving top performance on the MOT '15, '16, '17 and '20 Challenges among state-of-the-art online approaches. The code is available at \href{https://github.com/IPapakis/GCNNMatch}{this https URL}
\end{abstract}

\section{INTRODUCTION}

Multi-object tracking (MOT)  is  a widely studied computer vision problem of tracking multiple objects across video frames, that has several applications including autonomous vehicles, robot navigation, medical imaging, and visual surveillance \cite{ciaparrone2020deep}. 
One of the major paradigms in MOT is the tracking-by-detection paradigm, where an object detector is first used to extract object locations at each frame separately, followed by a tracker which associates detected objects across frames. The goal of the tracker is to solve the bipartite graph \emph{matching problem}, where every object instance in a past frame is associated to at most one object instance in the current frame, using pair-wise object affinities. There are two variants of the matching problem considered in MOT: online matching, where objects are associated only using past frames, and offline matching, where information from both past and future frames are used to track a given object. In this work, we only focus on the MOT problem involving online matching.

One of the conventional approaches in online matching is to learn appearance similarity functions among pairs of objects across consecutive frames through the use of Siamese Convolutional Neural Networks (CNN) architectures during training, e.g., using pairwise loss \cite{lee2018multiple,leal2016learning} and triplet loss\cite{wang2019towards}. However, these approaches treat feature extraction and object association as two isolated tasks and only deal with the optimization aspect of object association during testing using traditional algorithms such as the Hungarian \cite{kuhn1955hungarian} that leads to inferior accuracy. Another limitation is that these methods do not take into account the relative locations of objects during feature learning.

Recently there have been attempts to merge the feature extraction and object association tasks using Graph Neural Networks (GNN) \cite{wu2020comprehensive}, that have achieved state-of-the-art performance \cite{braso2019learning} on benchmark MOT problems. These approaches take advantage of the graph nature of the problem by using CNN to learn features and GNN to associate objects. By embedding appearance and geometric information into the graph structure, these approaches allow object features to be learned while taking into account object interactions in the network. Previous MOT approaches based on GNN \cite{inproceedings,li2020graph,braso2019learning,jiang2019graph} have attempted to satisfy bipartite one-to-one matching constraints using loss functions such as cross-entropy and softmax loss in an end-to-end architecture. However, there is room for improvement in the performance of such approaches \cite{inproceedings,li2020graph,jiang2019graph}.

\begin{figure}[t]
\begin{center}
   \includegraphics[width=1\linewidth]{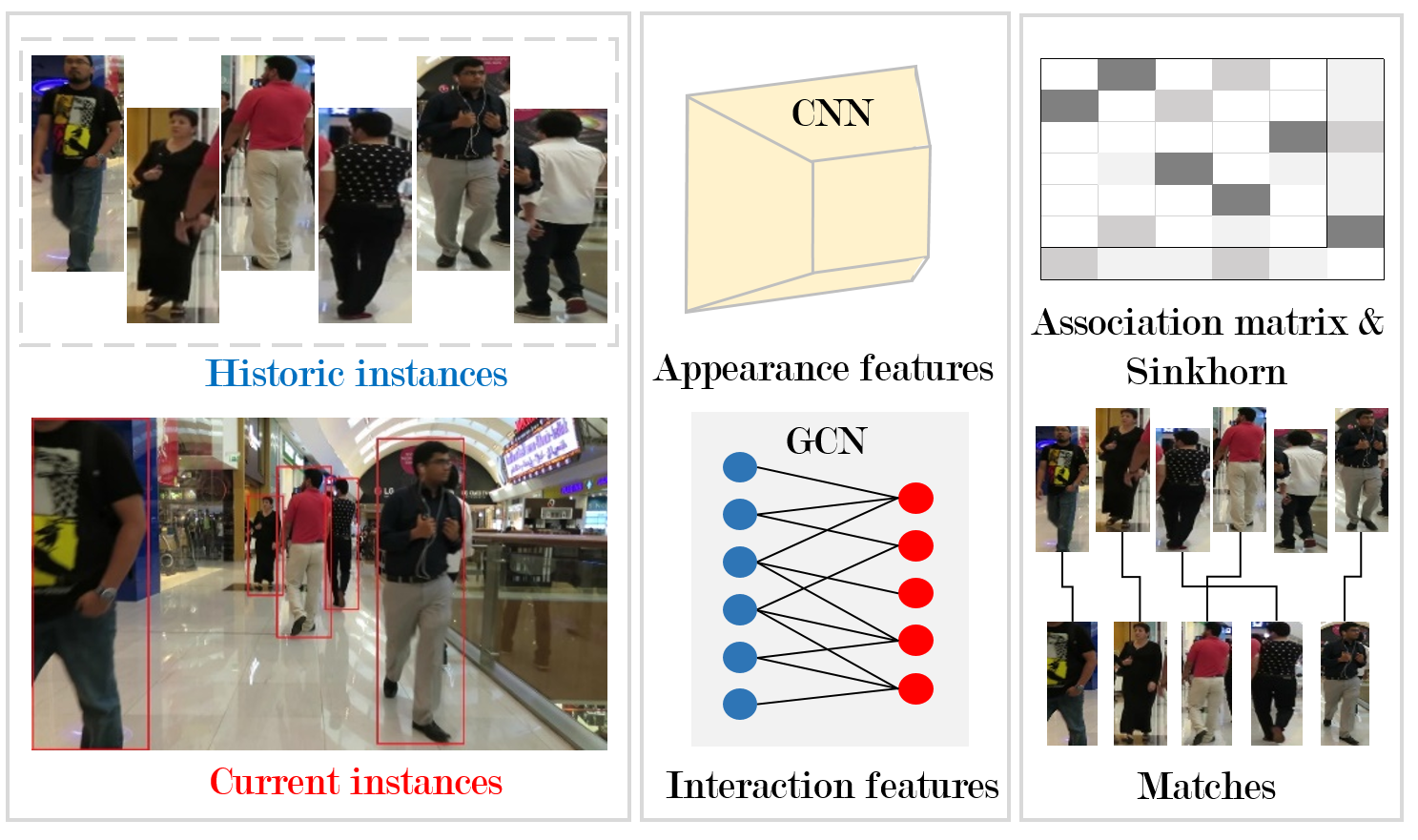}
   \caption{Illustration of the main components of our proposed approach. Historic object instances are matched with current frame detections, allowing objects to enter and exit the scene. Appearance and interaction features are used to produce similarity scores and to derive the final association using the Sinkhorn algorithm.}
\label{fig:intro}
\end{center}
\vspace{-4ex}
\end{figure}

In this paper, our goal is two-fold. First, we aim to develop an architecture that can combine GNN and CNN with geometric features in an online tracking paradigm to achieve top results in MOT, which previous online and GNN based methods have not achieved. Secondly, this method attempts to demonstrate that embedding MOT constraints in learning can be achieved without the use of extra loss functions such as softmax, but through an iterative process which can be more accurate. We propose a CNN and Graph Convolutional Neural Network (GCNN) based approach for MOT, depicted in Figure \ref{fig:intro}, to accurately solve the online matching problem subject to constraints specific to the MOT task. In our proposed approach, we model each object as a tracklet and feasible connections between tracklets from previous frames and new detections at the current frame form the edges of the graph. A CNN extracts appearance features from the last observed instance of each of the tracklets and a GCNN updates these features through the interaction of the nodes (tracklets) based on their connectivity. Finally, a Sinkhorn based normalization is applied to enforce the MOT constraints of the bipartite matching problem. Here is a summary of our contributions:

\begin{itemize}
\item We propose an online tracking method based on graph convolutional neural networks that achieves top performance in comparison to existing online approaches on the MOT benchmark. 

\item In contrast to traditional MOT approaches that learn appearance features of every object separately using Siamese architectures, our proposed approach operates on an arbitrarily large neighborhood of objects, incorporating context information such as location and object sizes using GCNN.

\item While previous GNN based approaches use loss functions to satisfy bipartite matching constraints, we introduce a novel approach of using the Sinkhorn normalization to enforce those constraints.

\item In contrast to other GNN based approaches for MOT, we use the geometric information not only during graph edge construction but also during affinity computation, thus significantly improving accuracy.
\end{itemize}
 
 The remainder of the paper is organized as follows. Section \ref{sec:rel} describes related work. Section \ref{sec:proposed} describes our proposed approach. Section \ref{sec:results} describes our evaluation setup and experimental results, while Section \ref{sec:conclusions} provides concluding remarks.
 
\section{RELATED WORK}
\label{sec:rel}

\subsection{Multi-Object Tracking}
A majority of previous work in MOT is based on the paradigm of tracking-by-detection \cite{ciaparrone2020deep}, which comprises of three basic stages. In the first stage of \emph{detection}, objects are identified at every frame using bounding boxes. In the next stage of \emph{feature extraction}, feature extraction methods are applied on the detected objects to  extract appearance, motion and other interaction features of the objects, which are then used to compute similarity or affinity scores among object pairs. In the final stage of \emph{association}, an assignment problem is solved to match objects at previous frames with objects at the current frame.

For feature extraction, a number of methods have been introduced for appearance feature extraction, including deep learning methods such as Siamese Networks \cite{lavi2018survey, kim2016similarity, lee2018multiple}, auto-encoders \cite{feng2017using, ho2020unsupervised}, correlation filters \cite{yang2019multi, kim2017multi}, feature pyramids \cite{lee2018multiple}, and spatial attention \cite{zhu2018online}. Motion extraction has also been an integral part of tracking and a number of methods have been developed utilizing Kalman Filters \cite{wojke2017simple}, optical flow \cite{wang2017online}, LSTM\cite{milan2017online}, among others. A number of methods have also been developed for computing pair-wise affinities. Common techniques include the use of metrics such as Intersection over Union and cosine similarity, LSTM variants (e.g., bi-directional \cite{yoon2019data}, bilinear \cite{kim2018multi}, and Siamese \cite{liang2018lstm}), and multi-layer perceptrons. The final task of association is commonly handled using a number of approaches such as the Hungarian algorithm \cite{wojke2017simple}, multiple hypothesis tracking \cite{chen2017enhancing}, dynamic programming \cite{ullah2018directed}, lifted multi-cut \cite{tang2017multiple} and reinforcement learning \cite{ren2018collaborative}.

Despite significant developments in the field of MOT, there is still a large margin for improving performance especially in terms of the number of objects that are accurately tracked throughout their lifespan in comparison to those which are not---important indicators of tracking performance. One of the limitations of aforementioned approaches for MOT  is that they perform feature learning without incorporating the interaction context of the features. As demonstrated in some recent approaches \cite{liugsm, braso2019learning}, incorporating the relative appearance and geometry of objects and allowing them to interact has the potential to create stronger matches and provide more robust associations, therefore increasing the number of accurately tracked objects and the overall tracking accuracy.

\subsection{Graph Neural Network based Tracking}
In an effort to incorporate object interactions during tracking and combine the steps of feature learning and matching, GNNs have recently been introduced for tracking. For example, a GCNN was used to update node features in \cite{inproceedings}, where the nodes are individual detections at every frame. After the GCNN updates, an adjacency matrix was computed using the cosine similarity of node embeddings, which was then used to assign detections to tracklets in an offline manner. Another approach proposed in  \cite{braso2019learning} uses Message Passing Networks to perform edge-based binary label propagations over the graph of detections. In another work by Jiang et al. \cite{jiang2019graph}, a method was proposed to  learn both an appearance model using two frames (similar to a Siamese network) and a geometry model using LSTM. The assignment task was solved using a GNN trained using three loss functions, for binary classification, for multi-class classification, and for birth or death of tracks. In Li et al. \cite{li2020graph}, the authors propose using two GNNs, one for learning appearance features and another for learning motion features.

While GNN based methods have a lot of promise for MOT, existing approaches have yet to become as accurate and robust as compared to other baselines, especially in the task of online tracking. We posit that one of the reasons for the limited accuracy of GNN based methods is that they satisfy the constraints of bipartite matching only using training losses or dedicated neural networks, while there may be other superior approaches for satisfying MOT constraints exactly during tracking and association. For example, as demonstrated in a related problem of point matching \cite{sarlin2020superglue}, the Sinkhorn algorithm is effective in ensuring constraint satisfaction and can be employed  both during training and testing, as compared to conventional association algorithms such as the Hungarian method that can only be invoked during testing. In contrast to existing GNN based methods, the use of the Sinkhorn algorithm during association is one of the key innovations of our proposed approach.

\begin{figure*}[t]
\begin{center}
\includegraphics[width=1\linewidth]{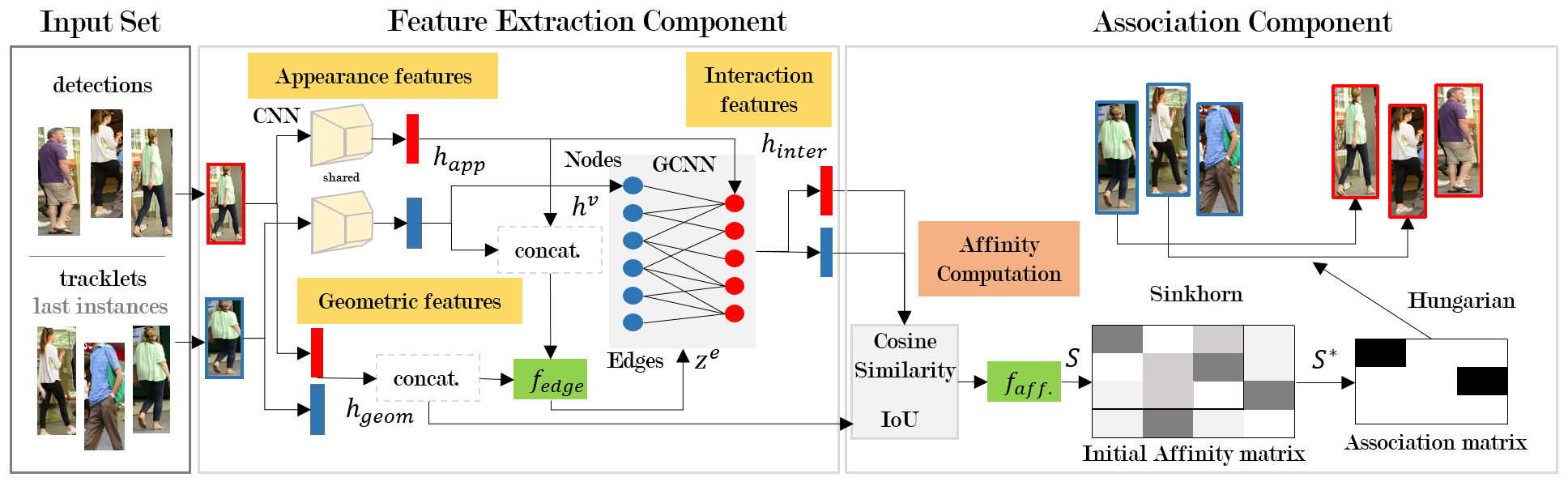}
\end{center}
   \caption{Overview of our proposed approach. Given an input set of bounding box images of tracklets and detections (nodes), we first extract appearance features $h_{app}$ using a CNN, which are used as node features $h^{v}$ in the GCNN. The appearance features, along with the geometric features $h_{geom}$, are then concatenated at node pairs and fed to a function $f_{edge}$ to compute edge features $z^{e}$. These node and edge features are used by GCNN to produce interaction features $h_{inter}$ at every node. Using cosine similarity and IoU, $f_{af\!finity}$ computes the similarity scores $S$ among pairs of tracklets and detections. The Sinkhorn algorithm then normalizes $S$ to match MOT constraints and produces the association matrix output $S^{*}$. During testing, the Hungarian algorithm is used for converting the values in $S^*$ to binary using a threshold.}
\label{fig:framework}
\end{figure*}

\section{PROPOSED APPROACH}
\label{sec:proposed}

\subsection{Problem Statement}

We are given a set of $N$ detections at a current frame $t$ as $D = \{D_1, D_2, \ldots D_N\}$ and a set of $M$ historic objects (or \emph{tracklets}) as ${T}=\{T_{1},T_2,\ldots,T_{M}\}$. We are also given bounding box images to represent the detections and tracklets. Note that while the bounding box images for detections correspond to the current frame $t$, the images for tracklets correspond to time-points when they were last observed in past frames. This means that this method utilizes the last instance from each tracklet, therefore $T_{m}=T^{last}_{m},\forall m\in[1,M]$. Furthermore, apart from the bounding box images, we also have information about the geometric features of every detection and tracklet, represented as a 4-length vector, $h_{geom} = \{\alpha,\beta,\gamma,\delta\}$, comprising of the bounding box center's horizontal position ($\alpha$), vertical position ($\beta$), box's width ($\gamma$), and height ($\delta$).

On the training set, we are given ground-truth labels $O$ for the association between detection $m$ and tracklet $n$ represented as $o_{m,n}$, which can either be 1 (match) or 0 (no match). Note that a tracklet can be associated to at most one detection, and it is possible for new detections to appear as well as existing tracklets to disappear at any frame. 
The goal of MOT then is to learn a  model that can predict the association between detection $m$ and tracklet $n$ as $s_{m,n}$ across all time-points. In other words, we want to learn the optimal association matrix $S^*$ such that: 
\begin{equation}
S^* = \text{argmax}_{S} \sum_{m=1}^M\sum_{n=1}^Ns_{m,n}o_{m,n}
\label{eq:1}
\end{equation}

The problem of learning $S^*$ can also be viewed as a bipartite graph matching problem, where the graph $G = (V,E)$ comprises of nodes $V = D \cup T$ and bipartite edges $E$ connecting a tracklet node $T_m \in T$ to a detection node $D_n \in D$ if there exists a match between $T_m$ and $D_n$ in $O$. The goal of MOT then is to learn a model to recover the adjacency matrix $S$ of the graph given the image appearance features and geometric features of the nodes as well as MOT matching constraints.


\subsection{Proposed Approach Overview}
Figure \ref{fig:framework} provides an overview of our proposed approach that comprises of two basic components. First, we extract features for tracklets and detections using a combination of CNN and GCNN. In particular, we use the CNN to extract \emph{appearance features} $h_{app}$  given the bounding box images of tracklets and detections. We also leverage the \emph{geometric features} $h_{geom}$ at every node, which are concatenated with $h_{app}$ at pairs of nodes to extract edge features $z^{e}$ using a fully connected neural network (FC-NN), $f_{edge}$. The extracted edge features $z^e$, along with node features $h^v = h_{app}$ are then fed to a GCNN to extract \emph{interaction features} $h_{inter}$ at every node. 

In the second component, we use the extracted features at nodes to compute affinities between every pair of tracklet $T_m$ and detection node  $D_n$ as follows. We first compute the cosine similarity between the interaction features $h_{inter}$ at $T_m$ and $D_n$. We then compute the intersection over union (IoU) of the bounding box areas represented by the geometric features $h_{geom}$ at $T_m$ and $D_n$. The cosine similarity and IoU are then fed to a FC-NN, $f_{af\!finity}$ to produce a real-valued score $s_{m,n}$ representing the affinity between $T_m$ and $D_n$. These affinity scores are then normalized across rows and columns using the Sinkhorn algorithm to satisfy the MOT constraints and produce the final association matrix $S^*$. During testing, the Hungarian algorithm is applied to binarize $S^*$ using a threshold to produce hard assignments between tracklets and detections. In the following, we provide brief descriptions of the two components of our proposed approach.


\subsection{Feature Extraction Component}

We use the bounding boxes for tracklets and detections available through publicly available detectors as the set of inputs for feature extraction. We first obtain the cropped image for every bounding box that are fed into a CNN architecture to extract appearance features $h_{app}$ of tracklets and detections, available as flat high-dimensional vectors. The conventional approach in MOT is to map such high-dimensional vectors to lower-dimensional embeddings using fully connected neural networks (FC-NN), which are then used for classification, re-identification, and many other tasks. However, by only using CNN and FC-NN, this approach does not incorporate the interaction effects between different objects (e.g., detections and tracklets) that are prevalent in MOT. To address this, we consider the goal of extracting interaction features at tracklets and detections using  a GCNN architecture instead of FC-NNs.

The inputs to our GCNN architecture consist of node and edge features, where the nodes comprise of tracklets and detections while edges denote bipartite matches between tracklets and detections. The node features $h^{v}$ at any node $v$ is simply the appearance features $h_{app}$ at $v$. To compute the edge features $z^{e}$ for a pair of tracklet  and detection nodes,  we first concatenate the appearance features $h_{app}$ and geometric features $h_{geom}$ at the pair of nodes and then send them to a FC-NN $f_{edge}$ to produce $z^{e} \in \mathrm{R}$.

Our GCNN architecture comprises of a number of hidden layers where at every layer $k$, the node and edge features produced at layer $k-1$ are non-linearly transformed using the neighborhood structure of the graph to produce node and  edge features at layer $k$, namely $H_{k}$ and $Z_{k}$, respectively. Note that at layer $0$, $H_{0} = H^{v}$ and $Z_{0} = Z^{e}$ are the input node and edge features, respectively. To understand the update operations at layer $k$, let us denote the adjacency matrix of the graph including self-edges at layer $k$ as $\tilde{Z_k} = Z_k + I$, where $I$ is an identity matrix. Further, let the degree of the adjacency matrix $\tilde{Z_k}$  be denoted by $\tau_k$. The node features are then updated at layer $k$ as:

\begin{equation}
H_{k}=\tau_{k}^{-1/2}\tilde{Z_{k}}\tau_{k}^{-1/2}H_{k-1}W_{k},
\end{equation}
where $W_k$ are learnable weights of the GCNN. Once $H_k$ has been updated, the edge features $Z_k(m,n)$ for an edge between nodes $v_m$ and $v_n$ are updated as:
\begin{equation}
Z_k(m,n)=\phi~(Z_{k-1}(m,n),H_{k}(v_m),H_k(v_n)),
\end{equation}
where $\phi$ is a FC-NN. The node features produced at the final layer are termed as the interaction features, $H_{inter}$.

\subsection{Association Component}\label{sec:3.3}
We use the interaction features $h_{inter}$ extracted by GCNN along with the geometric features $h_{geom}$ to compute the affinity of a tracklet $T_m$ to be associated with a detection $D_n$ using two simple metrics. First, we compute the cosine similarity between the $h_{inter}$ features at $T_m$ and $D_n$ to capture any interaction effects between the two objects discovered by the GCNN. Second, given the importance of the geometric features of $T_m$ and $D_n$ in determining their association affinity, we further compute the intersection over union (IoU) of  the bounding boxes of the two objects. This is different from existing GNN based approaches for MOT that only use the geometric information of objects during graph edge construction but not during affinity computation, thus making incomplete use of the information available in geometric features. Note that a higher value of IoU indicates a higher affinity score. We feed the cosine similarity score and IoU score to another FC-NN, $f_{af\!finity}$, that produces the affinity score, $s_{m,n}$.


Note that the affinity matrix S is constructed in such a way that each element represents the assignment score of tracklet $m$ to detection $n$. Since detections might not be associated with any tracklet and vice versa (denoting births and deaths of objects), we augment $S$ by adding a vector of rows and columns at the end of the matrix to produce a new $S$  of size $(M+1) \times (N+1)$. Further, note that the optimal $S$ is subject to the following MOT constraints:
\begin{eqnarray}
\sum^{N+1}_{n=1}s_{m,n} &= 1, ~~\forall m\in [1, \ldots, M]
\label{eq:2} \\
\sum^{M+1}_{m=1}s_{m,n} &= 1, ~~\forall n\in [1, \ldots, N]
\label{eq:3}
\end{eqnarray}
Further, at the last row  and column of $S$, we can further regularize $S$  (\cite{peyre2019computational, sarlin2020superglue}) using the following MOT constraints:
\begin{eqnarray}
\sum^{N+1}_{n=1}s_{m,n}&=N,  ~~m=M+1
\label{eq:4} \\
\sum^{M+1}_{m=1}s_{m,n}&=M,  ~~n=N+1
\label{eq:5}
\end{eqnarray}
We initialize $S$ using a default value of $s_{slack}\in R$. The conventional approach for satisfying MOT constraints (Equations \ref{eq:2}, \ref{eq:3}, \ref{eq:4}, and \ref{eq:5}) is to make use of specialized loss functions that can only be applied during training. In contrast, we leverage the Sinkhorn algorithm to automatically satisfy our MOT constraints both during training and testing, by iteratively normalizing the rows and columns of $S$ without the need for specialized loss functions. Each element $s_{m,n}$ is transformed using:
\begin{equation}s_{m,n}\leftarrow \lambda_{m} \frac{e^{l*s_{m,n}}}{\sum_{k=1}^{N+1} e^{l*a_{m,k}}}, ~ \lambda_{m} = \begin{cases} 1, & \mbox{if } m\in [1...M] \\ N, & \mbox{if } m=M+1 \end{cases}\end{equation}
\begin{equation}s_{m,n}\leftarrow \mu_{n} \frac{e^{l*s_{m,n}}}{\sum_{k=1}^{M+1} e^{l*s_{k,n}}}, ~ \mu_{n} = \begin{cases} 1, & \mbox{if } n\in [1...N] \\ M, & \mbox{if } n=N+1 \end{cases},
\end{equation}
where $l$ is a hyper-parameter representing the entropic regularization effect (larger value of $l$ generates greater separation in $S$). After a fixed number of iterations, the Sinkhorn algorithm produces the final association matrix $S^*$, where we drop the last row and column. Apart from satisfying the MOT constraints, an additional advantage of the Sinkhorn algorithm is that it is fully differentiable at every iteration. We can thus feed $S^*$ directly to the objective function of the end-to-end learning framework of our proposed approach, that involves minimizing the following weighted binary cross-entropy loss:
\begin{equation}
  \begin{aligned}[b]
L=-\frac{1}{MN}\times \Big( &\sum_{m,n} w*s_{m,n}\log(o_{m,n})+ \\&(1-s_{m,n})\log(1-o_{m,n})\Big)
    \end{aligned}
\end{equation}
where $w$ is a weight hyper-parameter to balance the imbalance among 1's and 0's in the ground-truth labels $O$. 
During testing, $S^*$ is first binarized using a cut-off threshold $s_{thres}$ and then the Hungarian method is applied over $S^*$ to perform hard assignments of 0 or 1. 

\section{EXPERIMENTAL ANALYSIS}
\label{sec:results}

\subsection{Datasets}
We evaluate our proposed approach on the publicly available MOT challenge  dataset \cite{milan2016mot16} that serves as a benchmark for comparing MOT performance of state-of-the-art methods using a standardized leader-board. We specifically focus on MOT challenge dataset since it includes annotations of detected objects including pedestrians in urban environments and has been widely used in the MOT community. MOT15 is the first version of the challenge, containing 11 train and 11 test sequences, each containing 145 to 1059 frames spanning diverse real-world environments. MOT16 contains 7 train and 7 test sequences, each containing 525 to 1,050 frames. The provided detections are obtained from DPM\cite{felzenszwalb2009object}. MOT17 is the same set as MOT16 but also contains detections from FASTER RCNN\cite{ren2015faster} and SDP\cite{yang2016exploit}. It also has more accurate ground truth and the majority of the methods are tested on that. MOT20 is the latest version of the challenge with 4 train and 4 test sequences and frames ranging from 429 to 2782. It contains crowded scenes of people in urban areas, however at the time of this writing, there are only a few entries in the challenge. From each of the provided datasets, the train set is split into training and validation sequences by holding off the last 150 frames of each video for validation. After training the model, we apply our proposed model on the test set using the online evaluation server \cite{milan2016mot16}. 

\subsection{Evaluation Metrics}
We consider standard metrics used in MOT literature and  reported on the MOT challenge leader-boards including the Multi-Object Tracking Accuracy (MOTA), Identity F1 score (IDF1), Mostly Tracked objects (MT, the ratio of ground-truth trajectories that are correctly predicted by at least 80\%), Mostly Lost objects (ML, the ratio of ground-truth objects that are correctly predicted at most 20\%), False Positives (FP), False Negatives (FN) and ID Switches (ID Sw.) \cite{milan2016mot16}.

\subsection{Implementation of Proposed Approach}
\paragraph{Network.} We used DenseNet-121 \cite{huang2017densely} as our choice of the CNN architecture with all fully connected layers at the end of the network replaced by GCNN. All activation functions used in the Network are ReLU. Also, all FC-NNs used as metric learner functions in our proposed approach consist of a simple architecture with no hidden layers. The GCNN consists of two hidden layers and in the output space of GCNN, $h_{inter}$, no activation function is used as the hidden layers contain sufficient non-linearity. The dimensionality of $h_{app}$ and $h_{inter}$ is 1,024 and 128, respectively, while the cropped bounding box image size is $150 \times 60$ pixels. The slack variable $s_{slack}$ was set at 0.2, while the regularization parameter $l$ was set at 5, and the number of Sinkhorn iterations set at 8. The weight hyper-parameter $w$ is set to 10 while $s_{thres}$ is set to 0.2. All  codes were developed in Pytorch\cite{NEURIPS2019_9015} and Pytorch-Geometric\cite{Fey/Lenssen/2019}.

\paragraph{Training Setup.}
We trained and tested our model on an Intel 2.6 GHz CPU cluster with NVIDIA TITAN RTX GPUs. The learning rate is set at $2\times 10^{-3}$ and regularization parameter $1\times 10^{-3}$ for Adam optimizer. Batch size is set to 12. Also, at each frame during training, we sample a random frame as previous frame going back up to 45 frames in order to provide more challenging matches. This introduces more cases of occlusion and significant appearance changes making our algorithm more robust during testing.

\paragraph{Baselines.}
We directly compare our approach with other \textit{online} approaches that are found in citable literature. Offline approaches such as \cite{braso2019learning} use detections from future frames as well, thus have a different scope.
Furthermore, public detections in MOT dataset are noisy and have many missing objects.  Since our approach is not improving the detections but performs only association of them, we pre-process the detections using Tracktor\cite{bergmann2019tracking}, similar to \cite{braso2019learning,liugsm}. Tracktor is considered as a baseline approach that can partially alleviate the problem of missing and false detections. Specifically, all detections with a confidence threshold less than 0.5 are ignored. For the remaining ones, Tracktor propagates a bounding box from the previous frame to the next by placing it into the same position and performing regression using the FRCNN regression head. Remaining detections are pruned using an NMS threshold of 0.8 and feasible matches are only created if the candidate objects fall within a pixel distance of 200. 

\begin{figure*}[h!]
\begin{center}
  \begin{subfigure}[b]{0.4\columnwidth}
    \includegraphics[width=4cm,height=3.5cm]{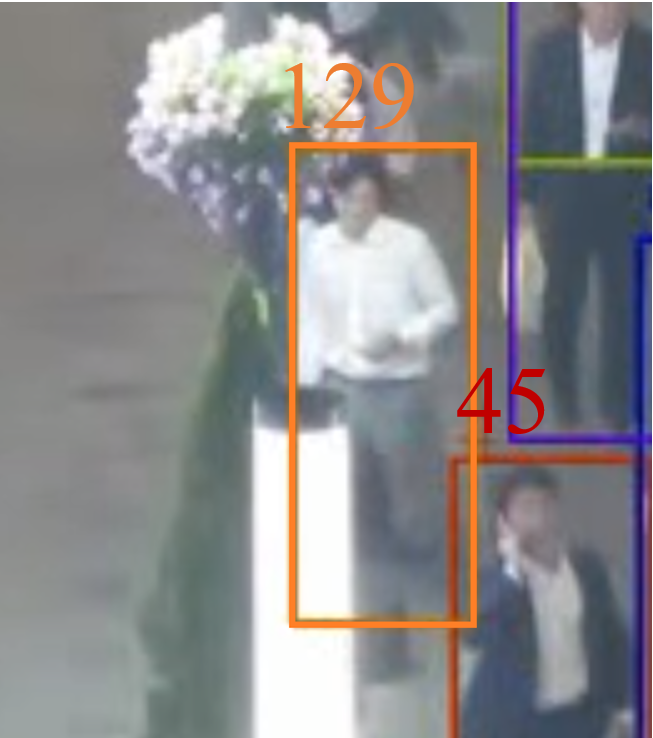}
    \caption{Video:  03, Fr: 542, Method: Ours}
    \label{fig:31}
  \end{subfigure}
  \hspace{2.25em} %
  \begin{subfigure}[b]{0.4\columnwidth}
    \includegraphics[width=4cm,height=3.5cm]{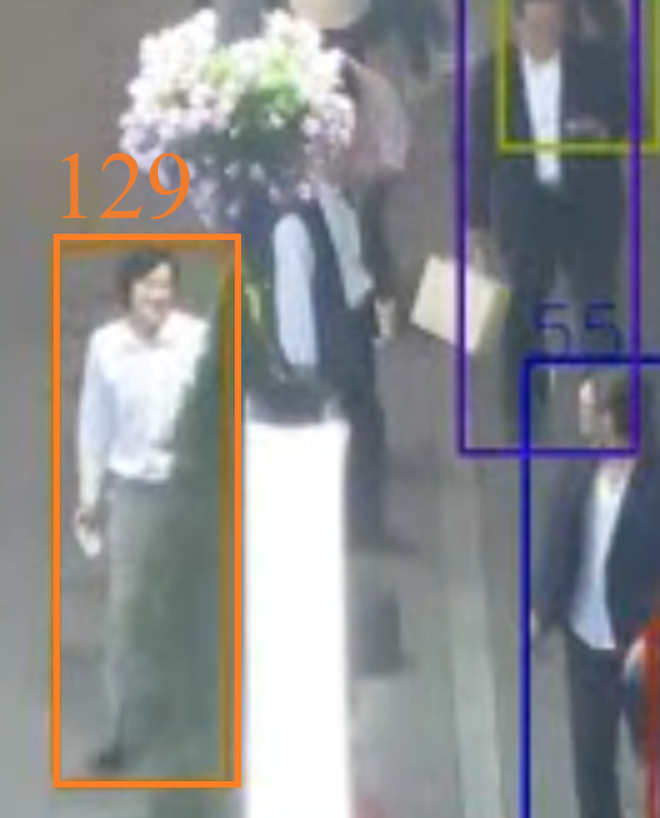}
    \caption{Video:  03, Fr: 570, Method: Ours}
    \label{fig:32}
  \end{subfigure}
  \hspace{2.25em} %
  \begin{subfigure}[b]{0.4\columnwidth}
    \includegraphics[width=4cm,height=3.5cm]{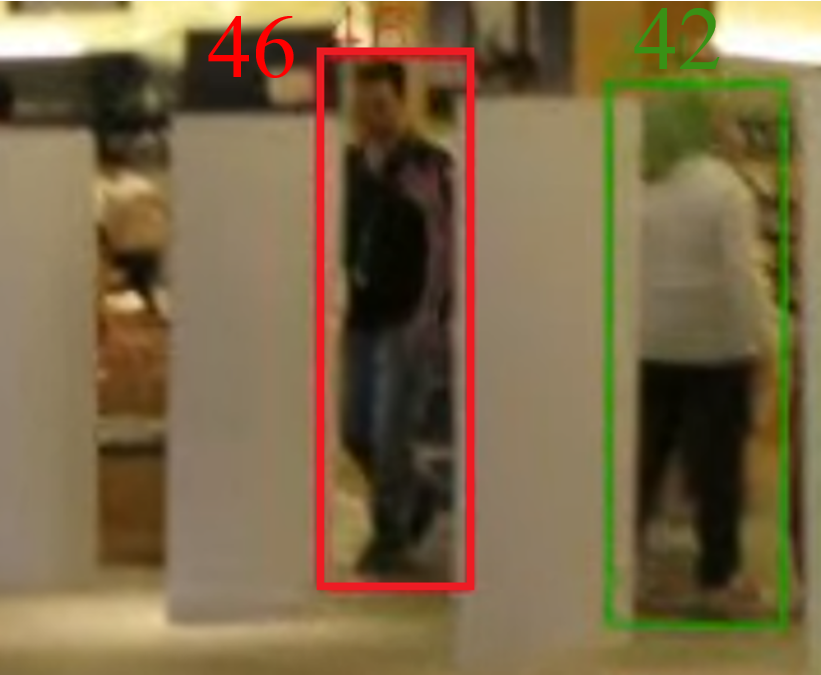}
    \caption{Video:  12, Fr: 556, Method: Ours}
    \label{fig:21}
  \end{subfigure}
  \hspace{2.25em} %
  \begin{subfigure}[b]{0.4\columnwidth}
    \includegraphics[width=4cm,height=3.5cm]{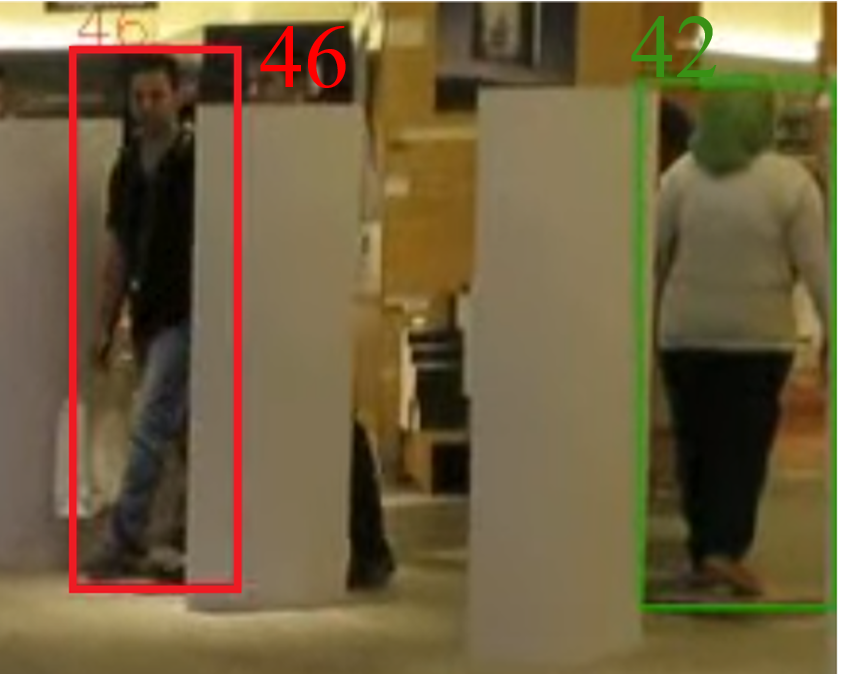}
    \caption{Video:  12, Fr: 583, Method: Ours}
    \label{fig:22}
  \end{subfigure}
\end{center}
\begin{center}
    \begin{subfigure}[b]{0.4\columnwidth}
    \includegraphics[width=4cm,height=3.5cm]{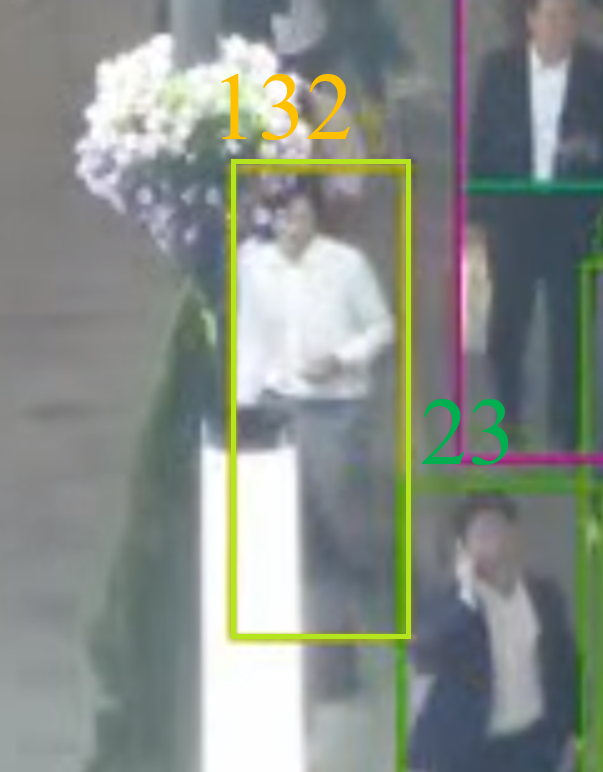}
    \caption{Video:  03, Fr: 542, Method: Tracktor-v2}
    \label{fig:33}
  \end{subfigure}
  \hspace{2.25em} %
  \begin{subfigure}[b]{0.4\columnwidth}
    \includegraphics[width=4cm,height=3.5cm]{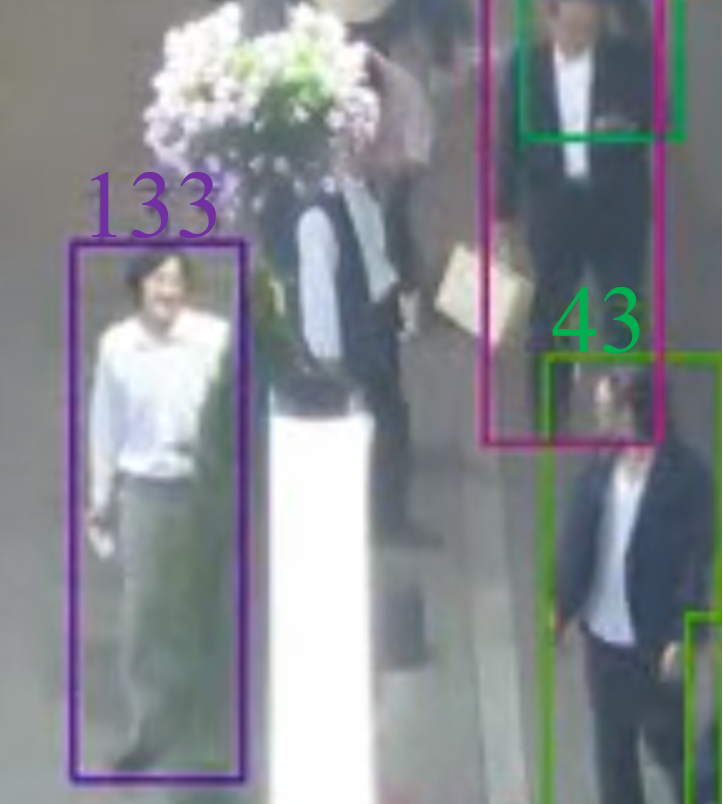}
    \caption{Video:  03, Fr: 570, Method: Tracktor-v2}
    \label{fig:34}
  \end{subfigure}
  \hspace{2.25em} %
\begin{subfigure}[b]{0.4\columnwidth}
    \includegraphics[width=4cm,height=3.5cm]{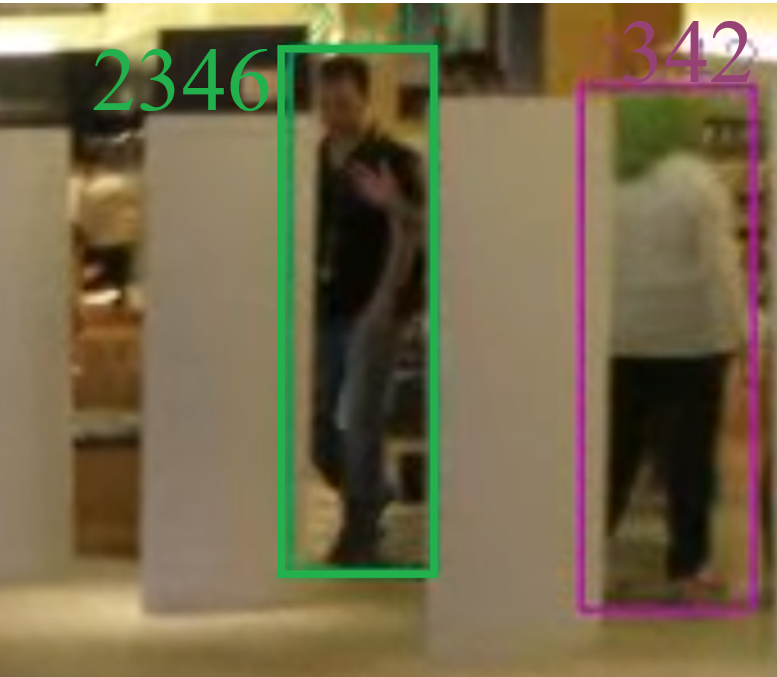}
    \caption{Video:  12, Fr: 556, Method: GSM-Tracktor}
    \label{fig:23}
  \end{subfigure}
  \hspace{2.25em} %
  \begin{subfigure}[b]{0.4\columnwidth}
    \includegraphics[width=4cm,height=3.5cm]{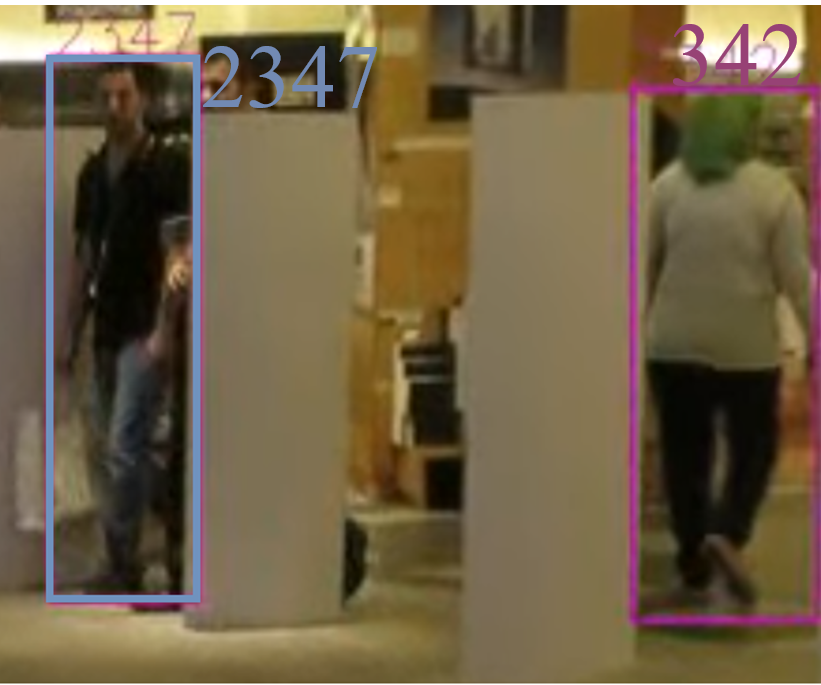}
    \caption{Video:  12, Fr: 583, Method:GSM-Tracktor}
    \label{fig:24}
    \end{subfigure}
     \caption{Qualitative analysis on MOT17-test set showcasing the accuracy in predicting object identities after occlusion. Each approach is shown for two frames. Each object has a colored box and an augmented number indicating its identity. In the first row, our method performance is shown for two frames (Fr) of two different videos (03 and 12). In the second row, a comparison is made using the same frames but with Tracktor-v2 \cite{bergmann2019tracking} and GSM-Tracktor \cite{liugsm}. Images obtained from \cite{MOTweb}.}.
\label{fig:qual1}
\end{center}
\vspace{-4ex}
\end{figure*}
\subsection{Results}

\begin{figure*}
\begin{center}
  \begin{subfigure}[b]{0.4\columnwidth}
    \includegraphics[width=4cm,height=3.5cm]{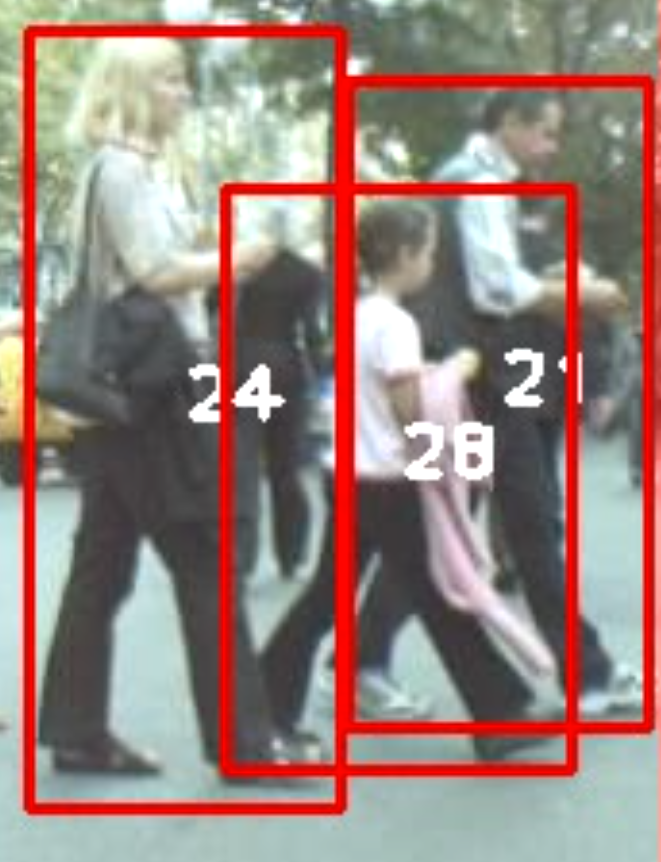}
    \caption{Video: 05, Fr.: 317, Ablation: FC-NN}
    \label{fig:qual2_a}
  \end{subfigure}
  \hspace{2.5em} %
  \begin{subfigure}[b]{0.4\columnwidth}
    \includegraphics[width=4cm,height=3.5cm]{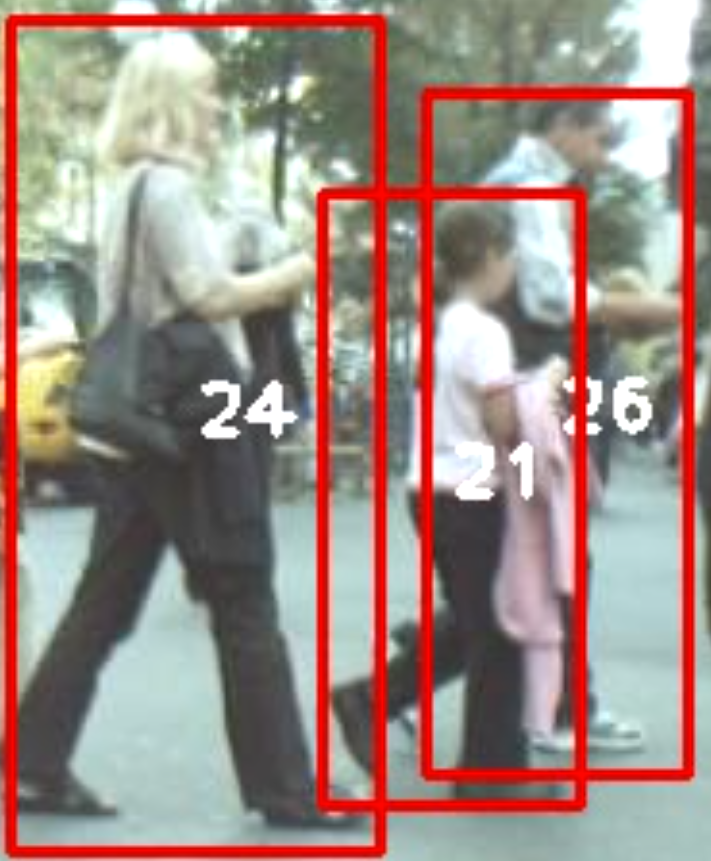}
    \caption{Video: 05, Fr.: 320, Ablation: FC-NN}
    \label{fig:qual2_b}
  \end{subfigure}
  \hspace{2.5em} %
\begin{subfigure}[b]{0.4\columnwidth}
    \includegraphics[width=4cm,height=3.5cm]{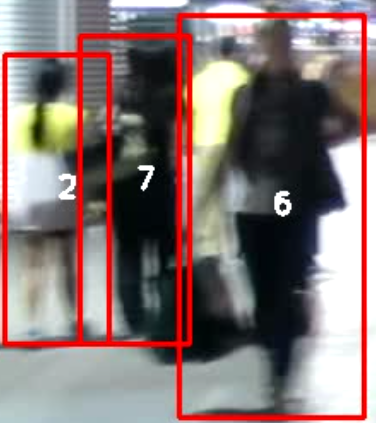}
    \caption{Video: 10, Fr.: 25, Ablation: Appear. only}
    \label{fig:qual2_e}
  \end{subfigure}
  \hspace{2.5em} %
  \begin{subfigure}[b]{0.4\columnwidth}
    \includegraphics[width=4cm,height=3.5cm]{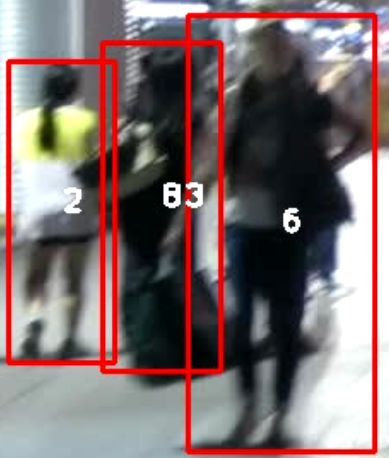}
    \caption{Video: 10, Fr.: 30, Ablation: Appear. only}
    \label{fig:qual2_f}
  \end{subfigure}
\end{center}
\begin{center}
  \begin{subfigure}[b]{0.4\columnwidth}
    \includegraphics[width=4cm,height=3.5cm]{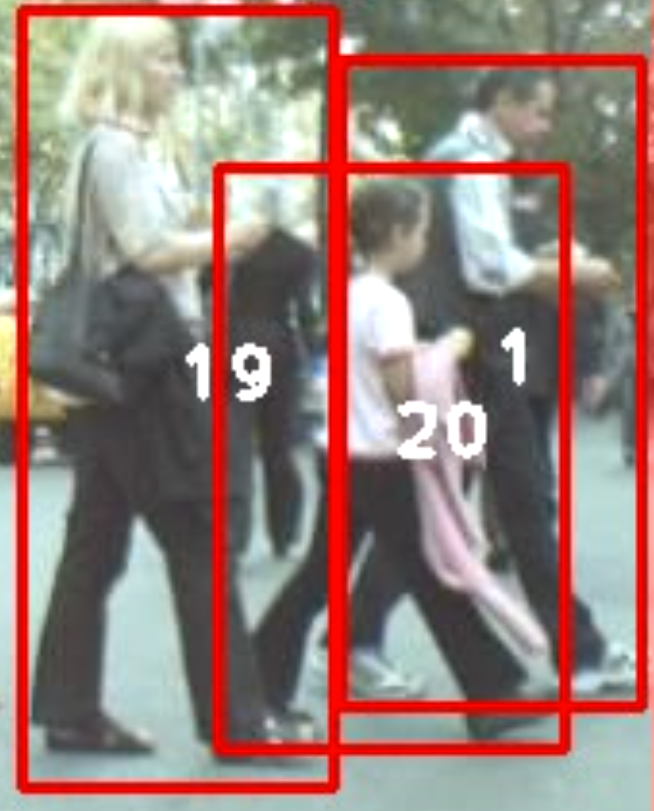}
    \caption{Video: 05, Fr.: 317, Ablation: GCNN}
    \label{fig:qual2_c}
  \end{subfigure}
  \hspace{2.5em} %
  \begin{subfigure}[b]{0.4\columnwidth}
    \includegraphics[width=4cm,height=3.5cm]{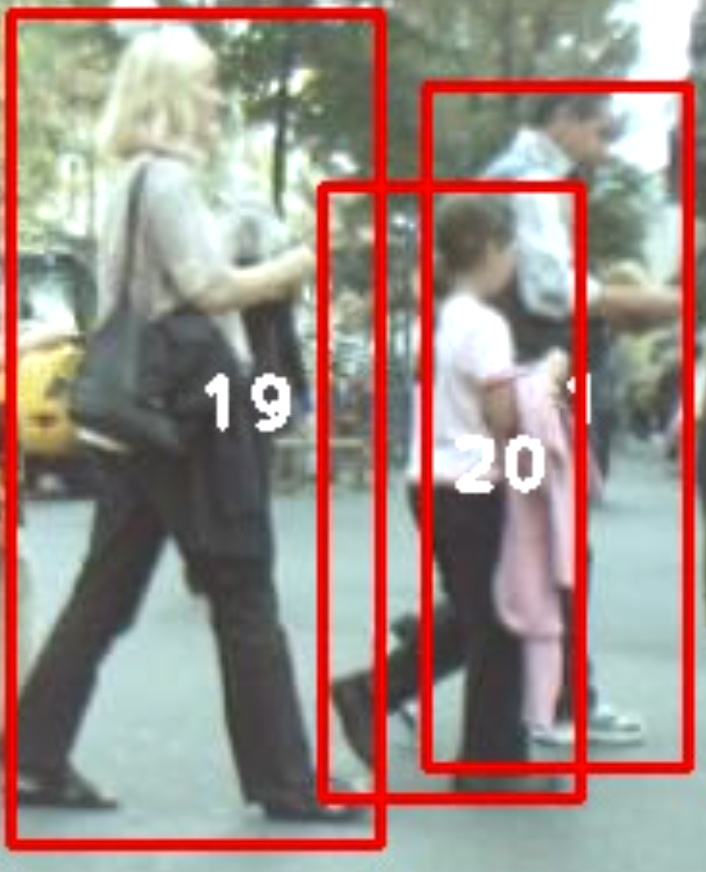}
    \caption{Video: 05, Fr.: 320, Ablation: GCNN}
    \label{fig:qual2_d}
  \end{subfigure}
  \hspace{2.5em} %
  \begin{subfigure}[b]{0.41\columnwidth}
    \includegraphics[width=4cm,height=3.5cm]{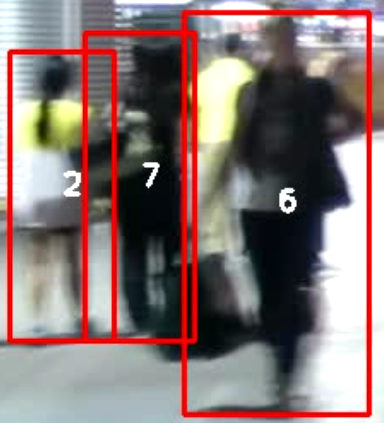}
    \caption{Video: 10, Fr.: 25, Ablation:Appear.\& Geom.}
    \label{fig:qual2_g}
  \end{subfigure}
  \hspace{2.5em} %
  \begin{subfigure}[b]{0.41\columnwidth}
    \includegraphics[width=4cm,height=3.5cm]{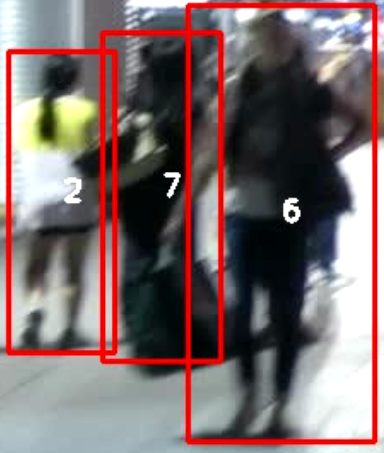}
    \caption{Video: 10, Fr.: 30, Ablation:Appear.\& Geom.}
    \label{fig:qual2_h}
  \end{subfigure}
   \caption{Qualitative analysis of performance on the MOT17-train set using different architectures during ablation study. Each object identity is illustrated using the drawn numbers inside each bounding box. In the first two columns, a comparison is performed using the GCNN and FC-NN based architectures. In the second two columns, the effect of the appearance and geometric features is examined.}
\label{fig:qual2}
\end{center}
\vspace{-2ex}
\end{figure*}

\begin{table}[h!]
\centering
 \caption{Comparison of our proposed approach with state-of-the-art supervised \textbf{online} trackers that use public detections on MOT Challenge Benchmark.}
\resizebox{\columnwidth}{!}{%
\Huge{
 \begin{tabular}{c c c c c c c c c} 
 \Xhline{2\arrayrulewidth}
  & & & &MOT 2015 \\
 \Xhline{2\arrayrulewidth}
  Method & MOTA 	$\uparrow$ & IDF1 $\uparrow$ & MT/ML $\uparrow$& MT $\uparrow$ & ML $\downarrow$  & FP $\downarrow$ & FN $\downarrow$ & ID Sw. $\downarrow$ \\ [0.5ex]
 \Xhline{2\arrayrulewidth}
 GCNNMatch (Ours) & 46.7 & 43.2 & 0.77 &157 & 203  & 6643 & 25311 & 1371 \\ 
 Tracktor-v2\cite{bergmann2019tracking} & 46.6 & 47.6 & 0.65&131 & 201& 4624& 26896& 1290
 \\ [1ex]
 TrctrD15\cite{bergmann2019tracking} & 44.1& 46.0& 0.65&124 & 192& 6085& 26917&
1868 \\ [1ex] 
 KCF\cite{chu2019online} &38.9 & 44.5& 0.53& 120& 227& 7321&29501 & 1440
 \\ [1ex] 
 STRN\cite{xu2019spatial} & 38.1& 46.6& 0.34& 83& 241& 5451& 31571&2665
 \\ [1ex] 
 AMIR15\cite{sadeghian2017tracking} & 37.6& 46.0& 0.59&114 &193 & 7933& 29397&
2024 \\ [1ex] 
 \Xhline{2\arrayrulewidth}
  & & & &MOT 2016 \\
  \Xhline{2\arrayrulewidth}
  Method & MOTA 	$\uparrow$ & IDF1 $\uparrow$ & MT/ML $\uparrow$& MT $\uparrow$ & ML $\downarrow$  & FP $\downarrow$ & FN $\downarrow$ & ID Sw. $\downarrow$ \\ [0.5ex]
 \Xhline{2\arrayrulewidth}
 GCNNMatch (Ours) & 57.2 & 55.0 & 0.67&174 &258 &3905 &73498 &559\\ 
 GSM-Tracktor\cite{liugsm} & 57.0 & 58.2 & 0.64&167 & 262 &4332 &73573 & 475 \\
 Tracktor-v2\cite{bergmann2019tracking} & 56.2 & 54.9 & 0.58&157 & 272 &2394 &76844 & 617\\
 TrctrD16\cite{xu2020train} & 54.8 &53.4 & 0.52&145 &281 &2955& 78.765 &645 \\
 MLT\cite{zhang2020multiplex} & 52.8 & 62.6 & 0.5& 160 &322 &5362 &80444 & 299  \\
 PV\cite{li2019multi} & 50.4 &50.8 & 0.38&113 &295 & 2600 &86780 &1061  \\
 GNMOT\cite{li2020graph} & 47.7 & 43.2 & 0.46& 120 &260 &9518 &83875 & 1907 \\[1ex]
 \Xhline{2\arrayrulewidth}
  & & & &MOT 2017 \\
 \Xhline{2\arrayrulewidth}
  Method & MOTA 	$\uparrow$ & IDF1 $\uparrow$ & MT/ML $\uparrow$& MT $\uparrow$ & ML $\downarrow$  & FP $\downarrow$ & FN $\downarrow$ & ID Sw. $\downarrow$ \\ [0.5ex]
 \Xhline{2\arrayrulewidth}
 GCNNMatch (Ours) & 57.3 & 56.3 & 0.67 &575 & 787  & 14100 & 225042 & 1911 \\
  GSM-Tracktor\cite{liugsm} & 56.4 & 57.8 & 0.64 & 523 & 813  & 14379 & 230174 & 1485  \\
 Tracktor-v2\cite{bergmann2019tracking} & 56.3& 55.1 & 0.6 & 498 &831  &8866 &235449 &1987 \\
 TrctrD\cite{xu2020train} & 53.7 &53.8 & 0.53 &458 &861  &11731 &247447 &1947 \\
 LSST\cite{feng2019multi} & 52.7 &57.9 & 0.49 &421 &863  &22512 &241936 & 2167  \\
 FAMNet\cite{chu2019famnet} & 52.0 & 48.7 & 0.57 &  450 & 787 &14138 & 253616 & 3072  \\
 YOONKJ\cite{yoon2020oneshotda} & 51.4 &54.0 &0.57 &500 &878 &29051 &243202 &2118 \\
 STRN\cite{xu2019spatial} & 50.9 &56.0 &0.56 &446 &797 &25295 &249365 &2397 \\
 MOTDT\cite{deeplylearned} & 50.9 &52.7 &0.49 &413 &841 &24069 &250768 &2474 \\
 GNMOT\cite{li2020graph} & 50.2 & 47.0 & 0.59 & 454 & 760 &29316 &246200 & 5273 \\
 MTDF\cite{fu2019multi}& 49.6 &45.2 &0.57 &444 &779 &37124 &241768 &5567\\
 DASOT\cite{chu2020dasot} & 49.5 &51.8 &0.59 &481 &814 &33640 &247370 &4142\\
 EDA-GNN\cite{jiang2019graph} & 45.5 &40.5 &0.39 &368 &955 &25685 &277663 &4091 \\[1ex] 
 \Xhline{2\arrayrulewidth}
  & & & &MOT 2020 \\
 \Xhline{2\arrayrulewidth}
  Method & MOTA 	$\uparrow$ & IDF1 $\uparrow$ & MT/ML $\uparrow$& MT $\uparrow$ & ML $\downarrow$  & FP $\downarrow$ & FN $\downarrow$ & ID Sw. $\downarrow$ \\ [0.5ex]
 \Xhline{2\arrayrulewidth}
GCNNMatch (Ours) & 54.5& 49.0& 1.28& 407& 317& 9522& 223611& 2456\\ 
Tracktor-v2\cite{bergmann2019tracking} & 52.6& 52.7& 1.1& 365& 331& 6930& 236680& 4374\\ [1ex]
 \Xhline{2\arrayrulewidth}
 \end{tabular}
 }
 }
 \label{sota}
\end{table}

\subsubsection{Benchmark Evaluation}
Table \ref{sota} shows a comparison of the performance of our proposed approach with that of top-performing online approaches on MOT leader-board. Our comparison has separate scopes. We first include GSM-Tracktor\cite{liugsm} which improves detections using Tracktor-v2. At the same time, the proposed approach is tested against other top approaches such as TrctrD\cite{xu2020train}. We further include online approaches that are based on GNN, GNMOT\cite{li2020graph} and EDA-GNN\cite{jiang2019graph}.

On the MOT15, our method achieves the highest MOTA of 46.7\%, marginally higher than Tracktor-v2 which is at 46.6\% but achieves a much higher MT/ML ratio of 0.77 contrary to the 0.65 of Tracktor. At the same time it surpasses methods such as TrctrD15\cite{xu2020train}, KCF\cite{chu2019online}, STRN\cite{xu2019spatial} and AMIR15\cite{sadeghian2017tracking} which range from 44.1\% to 37.6\% MOTA.

On the MOT16, it achieves 57.2\% and surpasses GSM-Tracktor and Tracktor-v2 by 0.2\% and 1\% respectively, while the rest of approaches including the Graph Neural Network Based GNMOT achieve only 54.8\% to 47.7\% MOTA.

On the MOT17, our method achieves the highest 57.3\% MOTA, highest MT/ML ratio of 0.73, highest 575 MT objects and smallest 225042 FN while maintaing a low number of 1911 ID switches. It is also located in top three places in IDF1,FP and ID switches. The results show improvement over both Tracktor-based baselines with 0.9 and 1.0 \% improvement in MOTA and 10\% and 15\% improvement on the number of MT. It further reduces FN by up to 3\%. Our method is able to surpass both its baselines, the siamese based re-identification model of Tracktor and the Graph Similarity model of GSM-Tracktor\cite{liugsm}.

At the same time, our approach surpasses the other methods, TrctrD\cite{xu2020train}, LSST\cite{feng2019multi}, FAMNet\cite{chu2019famnet}, YOONKJ\cite{yoon2020oneshotda}, STRN\cite{xu2019spatial}, MTDF\cite{fu2019multi} and DASOT\cite{chu2020dasot}. These approaches range from 53.7\% to 52.0\% MOTA and only LSST performs better at IDF1 but shows low number of MT. As it can be observed, GNMOT\cite{li2020graph} which utilizes a dedicated Single Object Tracker to improve detections and uses a GNN for association is lower than our approach by 7.1\% MOTA while EDA-GNN\cite{jiang2019graph} is also lower by 11.8\%. The combination of Tracktor-based detections with our robust association method performs better than the other combinations of approaches. 

On the MOT20, our method surpasses the Tracktor-v2 baseline by 1.9\% in MOTA and achieves 18\% higher MT/ML ration while the ID switches are much lower by 43.8\%.

Figure \ref{fig:qual1} showcases  qualitative results on difficult "after occlusion" cases for the association with two other trackers, GSM-Tracktor and Tracktor-v2 on the MOT17. The colored boxes along with the numbers on the boxes indicate the identity (IDs) of the objects. In the first two columns of images, a man with a white shirt and grey trousers is occluded for a few frames and then re-appears. In this case, our method recovers the identity of the person (Figures \ref{fig:31} and \ref{fig:32}) while Tracktor-v2 gives a new ID to the person (Figures \ref{fig:33} and \ref{fig:34}). In the second two columns of images, a man with dark clothes is occluded for a few frames. Our method identifies the same person, as shown in figures \ref{fig:21} and \ref{fig:22}, while GSM-Trackor identifies him as a new person (Figures \ref{fig:23} and \ref{fig:24}).

\subsubsection{Ablation Studies}
To understand the importance of the individual components of our proposed approach, we perform a series of studies using ablations of our complete model. We report the performance of these ablations on all videos in the MOT17 validation set, following a similar practise to \cite{li2020graph}.

In the first line of ablation studies, we evaluate the importance of using a GCNN instead of using an FC-NN that is typically used in traditional MOT methods. 
We also evaluate the importance of the Sinkhorn algorithm to satisfy the constraints of bipartite graph matching. 
It can be seen in Table \ref{ablation1} that applying the GCNN instead of the FC-NN produces an increase of 1.4\% MOTA, 3.8\% IDF1, while reducing ID switches by 37. We also adopt the Softmax loss proposed in \cite{weng2020gnn3dmot,jiang2019graph}, which yields 63.9\% MOTA and 72.4\% IDF1. On the other hand, using both GCNN and Sinkhorn produces 64.5\% MOTA and 76\% IDF1, increasing by 0.6\% the MOTA and 3.6\% the IDF1 in comparison to the Softmax Loss. This demonstrates the value of using GCNN along with Sinkhorn in our proposed approach.

In a second line of studies, we evaluated the importance of using the IoU metric to capture geometric features during affinity computation, in contrast to only using the geometric features for edge construction in GCNN. Table \ref{ablation2} shows the results of this study where the ``appearance only'' ablation corresponds to only using the cosine similarity in $h_{inter}$, while ``appearance + geometry'' correspponds to using both cosine and IoU. It can be seen that using IoU leads to an increase of 8.6\% in MOTA and an increase of 5.6\% in IDF1. This indicates that ignoring IoU leads to a weaker affinity score.

Finally, table \ref{ablation3} illustrates the importance of different number of layers used in the GCNN. Traditionally, GNNs do not require a large number of layers as  CNNs. It is shown that just using 2-layers produces higher MOTA and IDF1 than using 1 or 3 layers. Specifically, there is an increase of 0.4\% in comparison to 3 layers and an increase of 1.5\% in comparison to 1 layer.

Figure \ref{fig:qual2} provides a visual analysis of some ablation studies. The first two columns of images in Figure \ref{fig:qual2} compare the effect of using GCNN instead of FC-NN in our proposed approach. As the two people on the right become more occluded, an identity switch occurs for the case of ID:28 and ID:21 in Figures \ref{fig:qual2_a} and \ref{fig:qual2_b} when using FC-NN. On the other hand, Figures \ref{fig:qual2_c} and \ref{fig:qual2_d} show that by using GCNN  to capture interaction features, we obtain correct IDs despite the overlaps of the two boxes. In the next two columns, in Figures \ref{fig:qual2_e} and \ref{fig:qual2_f}, only the appearance is used while in Figures \ref{fig:qual2_g} and \ref{fig:qual2_h}, both appearance and geometry are used. It is clear that under blurry and low brightness conditions, a tracker using only appearance features for affinity computation is susceptible to ID switches.



\begin{table}[h!]
\caption{Ablation study on the effect of using GCNN instead of FC-NN and the Sinkhorn algorithm.}
\centering
\Huge{
\resizebox{\columnwidth}{!}{%
 \begin{tabular}{c c c c c c c c c} 
 \Xhline{2\arrayrulewidth}
 Method & MOTA 	$\uparrow$ & IDF1 $\uparrow$ & MT $\uparrow$ & ML 	$\downarrow$ & Prcn $\uparrow$ & Rcll $\uparrow$ & ID Sw.	$\downarrow$  \\ [0.5ex]
 \Xhline{2\arrayrulewidth}
  FC-NN \& Sinkhorn & 62.9 & 71.8 & 97 & 54 & 97.5 & 65 & 73 
  \\
 GCNN \& No Sinkhorn & 64.3 & 75.6 & 98 &54 &97.9 &65.9 &36\\
 GCNN \& Softmax Loss &  63.9&72.4&91&54&99.3&64.8&84\\
 GCNN \& Sinkhorn & \textbf{64.5} & \textbf{76} & \textbf{99} & \textbf{54} & \textbf{98.1} & \textbf{66} &\textbf{27}\\
[1ex] 
 \Xhline{2\arrayrulewidth}
 \end{tabular}
 }
 }
 \label{ablation1}
\end{table}
\begin{table}[h!]
\caption{Ablation study on the effect of using IoU (geometric features) during affinity computation.}
\centering
\Huge{
\resizebox{\columnwidth}{!}{%
 \begin{tabular}{c c c c c c c c c} 
 \Xhline{2\arrayrulewidth}
 Method & MOTA 	$\uparrow$ & IDF1 $\uparrow$ & MT $\uparrow$ & ML 	$\downarrow$ & FP $\downarrow$ & FN $\downarrow$ & ID Sw.	$\downarrow$  \\ [0.5ex]
 \Xhline{2\arrayrulewidth}
 Appear. Only & 55.9 & 70.4 & 87 & 69 & 96.8 &57.9 &23\\ 
 Appear.+Geom. & \textbf{64.5} & \textbf{76} & \textbf{99} & \textbf{54} & \textbf{98.1} & \textbf{66} &\textbf{27}\\[1ex] 
 \Xhline{2\arrayrulewidth}
 \end{tabular}
 }
 }
 \label{ablation2}
\end{table}

\begin{table}[h!]
 \caption{Ablation study on the effect of the number of layers used in the GCNN.}
\centering
\resizebox{\columnwidth}{!}{%
\Huge{
 \begin{tabular}{c c c c c c c c c} 
 \Xhline{2\arrayrulewidth}
 Method & MOTA 	$\uparrow$ & IDF1 $\uparrow$ & MT $\uparrow$ & ML 	$\downarrow$ & FP $\downarrow$ & FN $\downarrow$ & ID Sw.	$\downarrow$  \\ [0.5ex]
 \Xhline{2\arrayrulewidth}
 1-layer GCNN & 63.0 & 73.9 & 95 & 54 & \textbf{64.4} & 98.2 & 34 \\ 
 2-layers GCNN & \textbf{64.5} & \textbf{76} & \textbf{99} & \textbf{54} & \textbf{98.1} & \textbf{66} &\textbf{27} \\ 
 3-layers GCNN &  64.1& 74& 98 & 54 &98.3 &65.5 &32\\[1ex] 
 \Xhline{2\arrayrulewidth}
 \end{tabular}
 }
 }
 \label{ablation3}
\end{table}

\section{CONCLUSIONS}
\label{sec:conclusions}
In this paper, we have developed a novel method to handle online data association for Multi-Object Tracking. We have shown that using Graph Convolutional Neural Networks on top of Convolutional based features can achieve state-of-the-art tracking accuracy. A key innovation of our approach is to use a differentiable method, the Sinkhorn algorithm, to guide the association in an end-to-end learning fashion. Experimental results demonstrate top performance of our approach on the MOT Benchmark. The proposed framework opens the avenue for further research pertaining to the use of Graph Neural Networks into feature extraction as well as involving association into the learning pipeline. Future work on this method could involve summarizing the historic appearance of each tracklet for more accurate association under noisy detections.

{\small
\bibliographystyle{ieee_fullname}
\bibliography{egbib}
}

\end{document}